\documentclass{Interspeech2024}

\usepackage{tabularx}
\usepackage{booktabs}
\usepackage{multirow}
\usepackage{ragged2e}

\usepackage[backend=bibtex,citestyle=numeric-comp,bibstyle=ieee,sorting=none,defernumbers=true,giveninits=true,doi=false,isbn=false,url=false,eprint=false,minbibnames=2,maxbibnames=5]{biblatex}
\usepackage{pifont}
\newcommand{\cmark}{\ding{51}}%
\newcommand{\xmark}{\ding{55}}%

\usepackage{boldline}

\newcolumntype{C}{>{\Centering\arraybackslash}X}

\newcolumntype{u}{>{\raggedright\hsize=.7\hsize}X}
\newcolumntype{t}{>{\Centering\hsize=.6\hsize}X}
\newcolumntype{s}{>{\Centering\hsize=.5\hsize}X}
\newcolumntype{o}{>{\Centering\hsize=.4\hsize}X}
\newcolumntype{k}{>{\Centering\hsize=.3\hsize}X}
\newcolumntype{y}{>{\Centering\hsize=.2\hsize}X}
\newcolumntype{z}{>{\Centering\hsize=.1\hsize}X}
\newcolumntype{v}{>{\raggedright\hsize=.6\hsize}X}
\newcolumntype{e}{>{\raggedright\hsize=.5\hsize}X}
\newcolumntype{j}{>{\raggedright\hsize=.35\hsize}X}
\newcolumntype{f}{>{\raggedright\hsize=.3\hsize}X}
\newcolumntype{h}{>{\raggedright\hsize=.2\hsize}X}
\newcolumntype{q}{>{\raggedright\hsize=.8\hsize}X}

\interspeechcameraready

\title{Dynamic Data Pruning for Automatic Speech Recognition}

\name{{Qiao} {Xiao}$^{1,*}$, Pingchuan Ma$^{2,3,*}$, {Adriana} {Fernandez-Lopez}$^{2}$, {Boqian} {Wu}$^{4,5}$, {Lu} {Yin}$^{6}$, Stavros Petridis$^{2,3}$, {Mykola} {Pechenizkiy}$^{1}$, Maja Pantic$^{2,3}$, {Decebal Constantin} {Mocanu}$^{1,5}$, {Shiwei} {Liu}$^{7}$
\thanks{$^{*}$Equal contribution.}
}

\address{
  $^1$Eindhoven University of Technology \,
  $^2$Meta AI, UK \,
  $^3$Imperial College London\\
  $^4$University of Twente \,
  $^5$University of Luxembourg\\
  $^6$University of Surrey \,
  $^7$University of Oxford
  }
\email{{q.xiao}@tue.nl,\{pingchuanma,afernandezlopez\}@meta.com}

\keywords{automatic speech recognition, data pruning, learning efficiency}

\usepackage{multirow}
\newcommand{\code}[1]{\texttt{#1}}

\begin{document}
\maketitle

\begin{abstract}
The recent success of Automatic Speech Recognition (ASR) is largely attributed to the ever-growing amount of training data. However, this trend has made model training prohibitively costly and imposed computational demands. While data pruning has been proposed to mitigate this issue by identifying a small subset of relevant data, its application in ASR has been barely explored, and existing works often entail significant overhead to achieve meaningful results. To fill this gap, this paper presents the first investigation of dynamic data pruning for ASR, finding that we can reach the full-data performance by dynamically selecting 70\% of data. Furthermore, we introduce Dynamic Data Pruning for ASR (DDP-ASR), which offers several fine-grained pruning granularities specifically tailored for speech-related datasets, going beyond the conventional pruning of entire time sequences.
Our intensive experiments show that DDP-ASR can save up to 1.6$\times$ training time with negligible performance loss. 

\end{abstract}

\section{Introduction}

In the speech domain, the increasingly larger training datasets have significantly contributed to remarkable performance gains~\cite{kaplan2020scaling, touvron2021training, bommasani2021opportunities}. However, it also poses substantial challenges to training with limited computational resources. 
Some prior works have revealed that not all training instances are equally important for model training~\cite{katharopoulos2018not, toneva2018empirical, coleman2020selection}. This has led to inspiring efforts to improve the training efficiency of neural networks by either eliminating redundant data or prioritizing training instances based on their informational complexity~\cite{baldock2021deep, mirzasoleiman2020coresetsnips, mindermann2022prioritized}. 
Many recent works have also proposed diverse data pruning approaches to enhance training efficiency across various domains, such as computer vision~\cite{paul2021deep, qin2024infobatch, abbas162semdedup, mirzasoleiman2020coresets} and natural language processing~\cite{xia2024less, marion2023less, gunasekar2023textbooks}. 

Despite the potential benefits of data pruning, it has received limited attention in the domain of Automatic Speech Recognition (ASR). 
In a recent study, Boris et al.~\cite{bergsma2023cluster} introduced a pruning approach to first group similar instances together through clustering to minimize the dataset size while maintaining its representative characteristics. This approach explores similarities in the multidimensional feature space of a pre-trained large audio model.  
However, it requires multiple trials to derive more precise representations before data pruning, leading to additional overhead costs. To address this issue, for the first time, we introduce \textit{Dynamic Data Pruning (DDP)}~\cite{raju2021accelerating,qin2024infobatch, he2023large} in the context of ASR. DDP is a recently emerged data-pruning technique where only a subset of data is sampled and fed into the model throughout the training process. 

Specifically, we begin by conducting a comprehensive investigation into DDP for ASR pre-training, using various pruning criteria. Our findings reveal an encouraging discovery: through the adoption of a curriculum learning strategy~\cite{bengio2009curriculum}, we are able to train an ASR model using only 70\% of the data while achieving performance on par with that of the full-data training approach. To further enhance the efficacy of data pruning in ASR, we delve into a series of finely-tuned granularities meticulously crafted for speech-related data pruning. These granularities encompass the removal of individual time points as well as segments of temporal chunks. Our results demonstrate that by selectively removing consecutive samples, we can further improve the data efficiency of ASR. These empirical investigations culminate in the development of a novel data pruning approach for ASR, which we term Dynamic Data Pruning for ASR (DDP-ASR).

DDP-ASR incorporates both instance-wise and fine-grained {time-wise} granularities, allowing for the removal of a significant portion of data while achieving substantial practical speedup. Our extensive experiments on Librispeech demonstrate that, with a mixture of rule-of-thumb pruning rates, DDP-ASR can deliver up to 1.6$\times$ training speedups, while maintaining comparable performance to that achieved with full data. Additionally, we investigate the model's temporal robustness when trained on pruned subsets, revealing that our approach also brings benefits of robustness to audio clips with low sampling rates. To the best of our knowledge, our work is the first attempt to explore dynamic data pruning for ASR with novel pruning granularities specifically tailored for speech-related data,
presenting new opportunities for enhancing the training efficiency of speech-related models.

\section{Methodology}

\begin{figure*}[!t]
    \centering
    \vskip -0.2in
    \includegraphics[width=0.85\textwidth, trim=0 0.5cm 0 0]{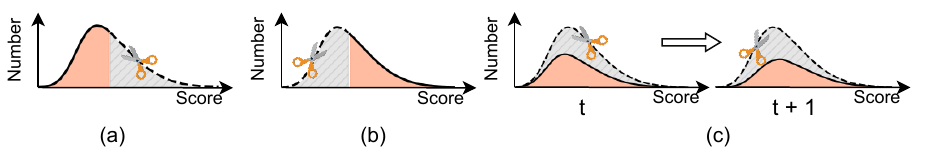}
    \vskip -0.05in
    \caption{Instance-wise pruning approaches: (a) \textit{Easy}: Instances with the lowest scores are selected. (b) \textit{Hard}: Instances with the highest scores are selected. (c) \textit{Easy2hard}: Following the essence of curriculum learning, models initially train on relatively easy instances and progressively shift focus to more challenging ones as the training progresses.}
    \label{fig:instance_prune}
    \vskip -0.2in
\end{figure*}

\subsection{Dynamic data pruning}
\vskip -0.05in
Given a dataset $\mathcal{D}=\left.\left\{z_i=\left(x_i, y_i\right)\right\}\right|_{i=1} ^{|\mathcal{D}|}$, a score $\mathcal{H}(z)$ is assigned to each instance $z$.  In each pruning cycle, instances are selectively removed according to the distribution of scores $\mathcal{H}$ and the pruning criterion $\mathcal{S}$. The reserved subset after data pruning is defined as:
\begin{equation}
\mathcal{X}^{kept}=\mathcal{S}\left(\mathcal{H}, \mathcal{X}, k\right)
\end{equation} 
where $k$ is the kept ratio and $\mathcal{X}$ corresponds to all input instances.
For \textit{static data pruning}~\cite{coleman2020selection,paul2021deep}, instances that satisfy a specific condition are permanently discarded before the training begins and will never be activated again. 

In contrast, dynamic data pruning enables the score 
$\mathcal{H}_t$ to be dynamically updated throughout training, ensuring that the coreset data can be more effectively adjusted based on the model's status at every $t$ epoch.
In this scenario, there is no reliance on pre-trained models, and intricate trials or runs are needed to acquire the pruning score before training~\cite{qin2024infobatch, raju2021accelerating, he2023large}.

\subsection{Dynamic data pruning for ASR}
\vskip -0.05in

\looseness -1 To facilitate training acceleration through dynamic data pruning for speech data, we introduce a novel data pruning method, referred to as DDP-ASR (Dynamic Data Pruning for ASR). DDP-ASR extends beyond the conventional instance-wise data pruning by incorporating fine-grained time-wise pruning strategies within each time sequence, thereby achieving a practical speedup.




\subsubsection{Instance-wise pruning}
\vskip -0.05in



Instance-wise pruning aims to remove entire audio sequences based on a given score $\mathcal{H}_t$. Several methods have been proposed to calculate the score of each instance, such as the loss values~\cite{qin2024infobatch} and uncertainty~\cite{he2023large}. The calculated score $\mathcal{H}_t$ is then used to determine which instances to preserve based on pruning criteria. For instance, the score distribution $\mathcal{H}_t$ can be used to identify and retain either {\textit{easy}} or {\textit{hard}} instances, where {\textit{easy}} instances are those with lower scores and {\textit{hard}} instances are those with higher scores.
%
In this study, we select the loss values $\mathcal{L}$ of each instance $z$ as the corresponding score, as these values can be obtained without extra cost during training and reflect the learning status of the instances. Moreover, its effectiveness has been verified in~\cite{cilimkovic2015neural, qin2024infobatch}.
Therefore, we explore a variety of instance-wise pruning methods tailored for ASR training:

\noindent\textbf{Easy.}\quad We prioritize the training of models on instances classified as ``easy'', which are identified by their lower scores, opting to exclude those with the highest scores. Figure~\ref{fig:instance_prune} (a) illustrates an example of this method.

\noindent\textbf{Hard.}\quad 
Conversely, we focus on incorporating ``hard'' instances for training, effectively sidelining those instances that are scored lower based on score distribution $\mathcal{H}_t$. Figure~\ref{fig:instance_prune} (b) provides an example for this method.

\noindent\textbf{Easy2hard.}\quad 
Inspired by curriculum learning strategies~\cite{wu2020curricula, soviany2022curriculum}, which train their models by progressively showing harder examples, we propose a novel selection strategy that dynamically schedules the presentation of instances to the model during training. Thus, at every checkpoint, $(1\,-\,\epsilon)\, k$ points with progressively increasing difficulty are kept, and $\epsilon \, k$ points are randomly selected from the remaining dataset. Here $\epsilon$ is used to strategically schedule the presentation of easy or hard instances to the model. It is worth noting that $\epsilon$ starts at 1 and gradually linearly decreases during training, effectively altering the selection strategy over time, as shown in Figure~\ref{fig:instance_prune} (c).

\subsubsection{Time-wise dropping}
\vskip -0.05in
\begin{figure}[!t]
    \centering
    \includegraphics[width=0.33\textwidth, trim=0.2cm 0.1cm 0 0]{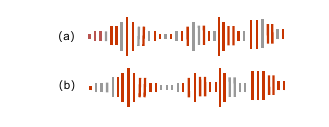}
    \vskip -0.15in
    \caption{A toy example comparing different time-wise pruning approaches: (a) Point Dropping, (b) Chunk Dropping, where signals highlighted in gray are pruned during training.}
    \label{fig:frame_drop}
    \vskip -0.2in
\end{figure}


\noindent\textbf{Point-wise dropping.} Inspired by CLIP~\cite{li2023scaling}, which removes a portion of image patches to yield a training speedup, we introduce \textit{point dropping}, a simple time-wise dropping strategy that removes individual data points within an instance to improve training speed.
This is done by randomly retaining $L$ audio samples within an instance that contains $T$ audio samples.




\noindent\textbf{Chunk-wise dropping.}\quad 
Compared to point dropping, \textit{chunk dropping} specifically targets the removal of chunks, each consisting of $n$ consecutive audio samples. For any given chunk $[t, t+n]$, $t$ is chosen from the range $[0, T-n)$, where $T$ is the input length of the instance. With a remaining length denoted as $L$, the method eliminates ${(T-L})/n$ chunks.

\section{Experimental Setup}
\noindent\textbf{Dataset.}\quad
In this study, we conduct experiments on two datasets: Librispeech~\cite{panayotov2015librispeech}, an audio corpus collected from audiobooks, and LRS3~\cite{afouras2018lrs3}, an audio-visual corpus from TED and TEDx talks. For Librispeech, we use ``train-clean-100'', ``train-clean-360'', and ``train-other-500'' subsets, totalling 960 hours of training data and evaluate our performance on the ``test-clean'' set with a total of 5.1 hours of audio. LRS3 consists of 439 hours of video clips, with~118\,516~(408 hours),~31\,982~(30 hours) and~1\,321 clips~(0.9 hours) in the pre-training, training-validation, and test sets, respectively. 


\noindent\textbf{Pre-processing.}\quad Following~\cite{DBLP:journals/corr/abs-2102-06657}, we take raw audio waveforms as input to the model and perform only z-normalisation per utterance before feeding it into the model.

\noindent\textbf{Data augmentation.}\quad
We only apply adaptive time masking~\cite{ma2022visual} to the raw audio stream. In particular, we choose a number of masks that is proportional to the utterance length and a maximum masking length of up to 0.4 seconds.

\noindent\textbf{Model architecture.}\quad
Rather than pursuing state-of-the-art performance, our primary goal is to investigate data pruning techniques in the domain of ASR. Accordingly, we adapt the open-source conformer-based architecture from~\cite{ma2023auto}. Our models comprises a 1D ResNet front-end (3.9 M parameters) to extract speech features from raw audio waveforms, followed by a conformer encoder (170.9 M parameters), a Transformer decoder (64.5 M parameters) and a projection CTC layer (3.9 M parameters), resulting in a total of 243.1 M parameters.

\noindent\textbf{Training details.}\quad
Following standard practices in ASR,  we train using a combination of CTC loss and Cross-Entropy loss. The model is trained for 75 epochs using the AdamW optimiser~\cite{loshchilov2017decoupled}. A cosine learning rate scheduler and a warm-up of 5 epochs are used, with the peak learning rate set to 0.001. We limit the duration of each training clip to no more than 16 seconds, and the maximum number of duration per batch is 64 seconds. All the models are trained with 32 A100 GPUs. For data pruning, we update the remaining subset every epoch. In the \code{Easy2hard} method, the proportion between the selected subset based on scores and a random one is set to 2\;:\;1 at the end of the training, which means that $\epsilon$ will linearly decay to 1\;/\;3. For time-wise dropping, we randomly drop samples up to the given dropping rate.

\section{Experimental Results}


\subsection{Performance for instance-wise pruning}
\vskip -0.05in
We evaluate the effectiveness of different instance-wise pruning methods for the Librispeech and LRS3 datasets. For a broader comparison, we include two additional pruning methods: (i) models trained with subsets randomly selected from the entire dataset in each pruning cycle, termed as \code{Random}, and (ii) models trained on a fixed subset initially chosen at random from the full dataset, referred to as \code{Static}.

Results of using different instance-wise pruning methods on the `test-clean'' set of Librispeech are shown in Table~\ref{tab:pruning_instance1}. We observe that for most pruning methods (namely, \code{Static}, \code{Easy} and \code{Random}, respectively), the performance is substantially impacted when more training instances are pruned. 
For example, when using 50\,\% easy training data (namely, \code{Easy}), a substantial increase of 1.7\,\% in Word Error Rate (WER) is observed. More hard instances likely tend to be removed, resulting in a relatively poor generalisation on the long sentences (More details analysis can be found in section~\ref{error_analysis}). The issue can be partly mitigated by training with the \code{Random} method, which avoids bias in the remaining instances. As a result, it narrows the performance gap to a mere 0.3\,\% in WER at a kept ratio of 50\,\%. 
A further closer performance gap to full data can be observed when using hard-related methods (namely, \code{Hard} and \code{Easy2hard}, respectively), which force the model to focus more on hard instances. 
Additionally, it is worth noting that using 70\,\% of the hard training instances can yield performance comparable to using the entire dataset, indicating considerable redundancy in LibriSpeech.

\begin{figure}[!tb]
    \centering
    \includegraphics[width=0.45\textwidth, trim=0 0.5cm 0.5cm 0]{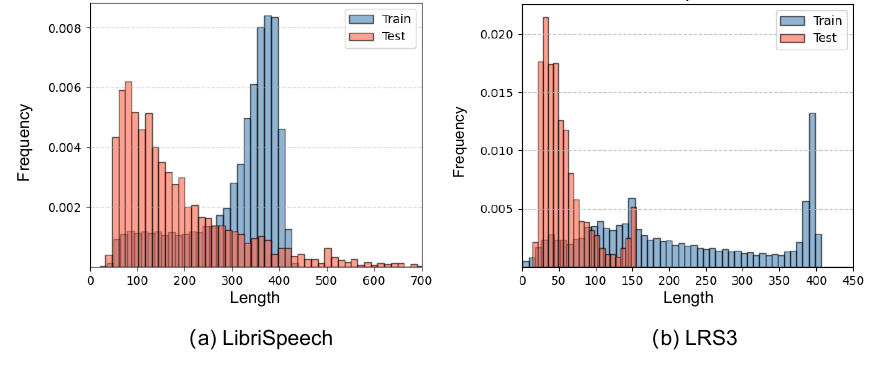}
    \vskip -0.02in
    \caption{The distribution of length on Librispeech and LRS3.}
    \label{fig:speech-length}
    \vskip -0.2in
\end{figure}

Results of using different instance-wise pruning methods on the test set of LRS3 are presented in Table~\ref{tab:pruning_instance2}. A similar trend as in the Librispeech experiments can be observed. The only exception is the results after using \code{Hard}, which consistently perform worse than the \code{Random} method. This might be due to a large discrepancy in the distribution of length between the training and test sets (as shown in Figure~\ref{fig:speech-length}). Specifically, concentrating on a subset of hard instances in the training set, which may not align well with the test set, can result in diminished test set performance. This is not the case for Librispeech, where length discrepancies are less noticeable.


\begin{table}[!t]
\centering
\renewcommand\arraystretch{1}
\tiny
\caption{WER [\%] ($\downarrow$) of our models with different pruning methods as a function of the kept ratio on the test set of Librispeech. The best results are bold for each kept ratio.}
\vskip -0.1in
\resizebox{1.0\linewidth}{!}{
\begin{tabular}{lccccc}
\toprule
\textbf{Method}/\textbf{Kept ratio [\%]}&\textbf{100} & \textbf{90} & \textbf{70} & \textbf{50} & \textbf{30} \\ 
\midrule
Static &2.58 &2.69 & 2.86 & 3.52 &4.59  \\ \midrule
Easy &2.58 &2.91 & 3.65 & 4.28 & 6.90   \\ \midrule
Random &2.58 & 2.59 & 2.66 & 2.88 & 3.17 \\  \midrule
Hard &2.58 &2.63 &2.57 & 2.76 & 3.13  \\ \midrule
Easy2hard &2.58 & \textbf{2.56} & \textbf{2.53} & \textbf{2.72} & \textbf{3.09} \\
\bottomrule
\end{tabular}
}
\label{tab:pruning_instance1}
\end{table}

\begin{table}[!t]
\renewcommand\arraystretch{1}
\tiny
\caption{WER [\%] ($\downarrow$) results of our ASR model with different pruning methods as a function of the kept ratio on the ``test-clean'' set of LRS3.}
\vskip -0.1in
\resizebox{1.0\linewidth}{!}{
\begin{tabular}{lccccc}
\toprule
\textbf{Method}/\textbf{Kept ratio [\%]}&\textbf{100} & \textbf{90} & \textbf{70} & \textbf{50} & \textbf{30} \\ 
\midrule
Static &2.10 &2.18 & 2.65 &3.45  &4.71  \\ \midrule
Easy &2.10 &2.27  & 2.27 & 5.03 & 15.17  \\ \midrule
Random &2.10 &2.12 & 2.12 & 2.29 & 2.60 \\  \midrule
Hard &2.10 &2.16 & 2.67 & 2.80 & 2.90  \\ \midrule
Easy2hard &2.10 &\textbf{2.08}  & \textbf{1.95}  & \textbf{2.17} & \textbf{2.54} \\
\bottomrule
\end{tabular}
}
\label{tab:pruning_instance2}
\end{table}

\begin{table}[!ht]
\centering
\renewcommand\arraystretch{1.0}
\scriptsize
\caption{WER [\%] ($\downarrow$) results of our models with different time-wise dropping methods as a function of the kept ratio on the ``test-clean'' set of LibriSpeech. The best results are bold for each time-wise kept ratio. ``k'' denotes the kept ratio.}
\vskip -0.1in
\resizebox{1.0\linewidth}{!}{
\begin{tabular}{ccccccc}
\toprule
\multirow{2}{*}{\textbf{Method}} &\multirow{2}{*}{\textbf{\shortstack{Instance-wise\\\textit{k} [\%]}}}&\multicolumn{5}{c}{\textbf{Time-wise \textit{k} [\%]}} \\
& & \textbf{100} & \textbf{90} & \textbf{70} & \textbf{50} & \textbf{30} \\ 
\midrule
Point & 70 &2.53 &2.65 &2.78 &2.80 &3.05 \\
\midrule
Chunk & 70 &2.53 &\textbf{2.59} &\textbf{2.59} &\textbf{2.66} &\textbf{2.79}\\
\bottomrule
\end{tabular}
}
\label{tab:frame-drop}
\vskip -0.2in
\end{table}

\subsection{Performance for time-wise dropping}
\vskip -0.05in
Table~\ref{tab:frame-drop} studies the impact of two time-wise dropping strategies (namely, \code{point} and \code{chunk}, respectively) by varying the dropping ratio on our proposed \code{Easy2hard} method. We observe that overall the performance gap between the model trained after time dropping and the baseline model without time dropping becomes increasingly larger as the dropping ratio increases. The performance degradation may be partially due to corrupted temporal dependencies, where the model relies on the precise order of input data for accurate predictions.
In particular, for point dropping, where at 30\% of the dropping ratio, it results in a 0.25\% increase of WER. Notably, chunk dropping, which involves removing consecutive samples in each chunk, can mitigate some of the performance declines. When applying a 30\% time-wise dropping ratio to the audio samples using this method, the impact on performance is minimal, with only a 0.06\% WER increase. Given that chunk dropping performs better, we use chunk dropping as our default setting for the time-wise dropping.

\subsection{Instance-wise pruning or time-wise dropping?}
\vskip -0.05in
We investigate in Table~\ref{tab:combination} the optimal strategy of combining both pruning methods under the same wall-clock training time. The results indicate that for models trained with a larger portion of data (more than 70\%), including time dropping results in a slight decrease in performance. Interestingly, we show that when the sampling rate is down-sampled from 16\,000\,Hz to 11\,025\,Hz, a closer performance gap for the model with time dropping can be observed compared with its counterpart, which indicates that a more robustness temporal dependency is learnt when time-wise dropping is applied. On the other hand, within the same training duration, combining instance- and time-wise data pruning for a smaller portion of training data leads to an observable reduction in WER, compared to models trained solely with instance-wise data pruning at a standard sampling rate (16\,000\,Hz). This suggests that the synergistic application of both pruning strategies is beneficial in scenarios with a very limited computational resource. In general, it is observed that models trained using time-wise dropping exhibit greater robustness across different sampling rates, especially at a low sampling rate.

\begin{table}[!t]
\renewcommand\arraystretch{1.05}
\normalsize
\centering
\caption{Impact of different sampling rates on the performance of Librispeech. ``k'' denotes the kept ratio.}
\vskip -0.1in
\resizebox{0.99\linewidth}{!}{
\begin{tabular}{ccccc} 
\toprule
\textbf{\shortstack{Instance\\\textit{k} [\%]}}&\textbf{\shortstack{Time\\\textit{k} [\%]}} &\textbf{\shortstack{Wall-clock time\\per epoch [min]}}  &\textbf{\shortstack{WER\\16KHz [\%]}} &\textbf{\shortstack{WER\\11KHz [\%]}}\\
\midrule
100 &100 & 13.2$\pm$0.2 &2.58 &9.77   \\ 
\midrule
70 &100 &9.8$\pm$0.2 &2.53 &10.56 \\ 
80 &80 &9.8$\pm$0.1 &2.56$_{\textcolor{red}{\uparrow 0.03}}$ &8.94$_{\textcolor{blue}{\downarrow 1.62}}$ \\
90 &60 &9.7$\pm$0.2 &2.61$_{\textcolor{red}{\uparrow 0.08}}$ &7.48$_{\textcolor{blue}{\downarrow 3.08}}$  \\
\midrule
30 &100 &4.2$\pm$0.1 & 3.09 &14.29 \\ 
40 &50 &4.2$\pm$0.1 &3.04$_{\textcolor{blue}{\downarrow 0.05}}$ &9.52$_{\textcolor{blue}{\downarrow 4.77}}$    \\ 
50 &25 &4.4$\pm$0.2 &2.99$_{\textcolor{blue}{\downarrow 0.10}}$&9.42$_{\textcolor{blue}{\downarrow 4.87}}$  \\
\bottomrule
\end{tabular}
}
\label{tab:combination}
\vskip -0.2in
\end{table}



\begin{table}[!h]
\renewcommand\arraystretch{.5}
\scriptsize
\centering
\caption{Time masking and dropping.  Instances are kept to 70\,\% of the whole dataset for all cases. We mask up to 40\,\% audio samples in chunks and drop up to 30\% of the audio samples.}
\vskip -0.1in
\resizebox{0.9\linewidth}{!}{
\begin{tabular}{ccccc} 
\toprule
\textbf{Time mask} &\textbf{Time drop}&\textbf{\shortstack{Wall-clock time\\per epoch [min]}}  & \textbf{WER [\%]}\\
\midrule
\xmark &\xmark &9.7$\pm$0.2 &3.11 \\ 
\midrule
\cmark &\xmark &9.8$\pm$0.2 &\textbf{2.53} \\ 
\midrule
\xmark &\cmark &\textbf{8.1$\pm$0.1} &2.78 \\ 
\midrule
\cmark &\cmark &\textbf{8.1$\pm$0.1} &{2.59} \\ 
\bottomrule
\end{tabular}
}
\label{tab:mask_vs_drop}
\vskip -0.2in
\end{table}

\subsection{Time masking versus time dropping}
Time dropping is implemented by eliminating data points from the training instances, unlike time masking, which sets consecutive samples to zero without altering the speed, as discussed in~\cite{DBLP:conf/interspeech/ParkCZCZCL19}. Table~\ref{tab:mask_vs_drop} shows the impact of the use of time masking and time dropping on the ``test-clean'' set of Librispeech. In particular, the use of time masking results in a 0.58\,\% reduction in WER. However, substituting time masking with time dropping leads to a 0.25\,\% increase in WER, alongside a significant reduction of ~17\,\% times. Interestingly, by integrating both time-masking and time-dropping approaches, it is possible to mitigate the performance decrease, achieving enhanced efficiency and comparable performance to the original model.


\subsection{Data scaling and speedup}
We expanded the training dataset from 960 hours to 3,494 hours by incorporating additional datasets such as LRS3, VoxCeleb2, and AVSpeech. The outcomes on the Librispeech dataset, displayed in Table~\ref{scaling}, indicate a marginal improvement in performance using our proposed data pruning method compared to the random pruning method. This demonstrates the effectiveness of our approach when applied to larger training datasets.

Table~\ref{epoch_steps} presents the performance comparison of our method with full data training under the same training time. In particular, when using an instance-wise kept ratio of 70\% with the \code{Easy2hard} method, which takes a similar training time to the model using full data trained for 56 epochs, we observe a further decrease in WER by 0.06\% compared to the model trained with the entire dataset for 56 epochs. Moreover, when combined with time-wise dropping, the training speed improves by 38\%, resulting in a comparable WER to that achieved with 75 epochs of full data training.


\begin{table}[!t]
\caption{Impact of the size of additional training data on the "test-clean" set of Librispeech. The additional data includes LRS3, VoxCeleb2, and AVSpeech, totaling 3\,494 hours.}
\vskip -0.1in
\renewcommand\arraystretch{1.1}
\small
\centering
\resizebox{0.99\linewidth}{!}{
\begin{tabular}{ccccc} 
\toprule
\textbf{\shortstack{Training\\ data}}&\textbf{\shortstack{Instance\\\textit{k} [\%]}}&\textbf{\shortstack{Time\\\textit{k} [\%]}} &\textbf{\shortstack{Wall-clock time\\per epoch [min]}} & \textbf{WER [\%]}\\
\midrule
\multirow{2}{*}{\vspace*{0pt}\shortstack{Random\\ Easy2hard}} & 50 &100 &25.5$\pm$0.5 &2.25   \\
&50 &100  &26.8$\pm$1.0 &\textbf{2.18}   \\
\bottomrule
\end{tabular}
}

\vskip -0.1in
\label{scaling}
\end{table}




\begin{table}[!t]
\caption{Impact of the number of training epochs on the Librispeech dataset. ``k'' denotes the kept ratio.}
\vskip -0.1in
\renewcommand\arraystretch{1.1}
\small
\resizebox{1.0\linewidth}{!}{
\begin{tabular}{ccccc} 
\toprule
\textbf{\shortstack{Instance-wise\\\textit{k} [\%]}}&\textbf{\shortstack{Time-wise\\\textit{k} [\%]}}&
\textbf{\shortstack{Training\\ epochs}}
&\textbf{\shortstack{Wall-clock time\\per epoch [min]}} & \textbf{WER [\%]}\\
\midrule
\multirow{2}{*}{\vspace*{0pt}\shortstack{100}} &\multirow{2}{*}{\vspace*{0pt}\shortstack{100}} &75 &\multirow{2}{*}{\vspace*{0pt}\shortstack{13.2$\pm$0.2}}
 &2.58   \\
\cmidrule{3-3}\cmidrule{5-5}
%
 &  &
56$_{\textcolor{blue}{\downarrow 25\,\%}}$& & 2.59   \\
\midrule
70 &100 &75&{9.8}$\pm${0.2}$_{\textcolor{blue}{\downarrow 25\,\%}}$ &\textbf{2.53}   \\
70 &70 &75&{8.1}$\pm${0.1}$_{\textcolor{blue}{\downarrow 38\,\%}}$ &{2.59}   \\
\bottomrule
\end{tabular}
}
\label{epoch_steps}
\end{table}

\subsection{Error analysis}
\begin{figure}[!tb]
\vskip -0.1in
    \centering
    \includegraphics[width=0.35\textwidth, trim=0 0.5cm 0.5cm 0]{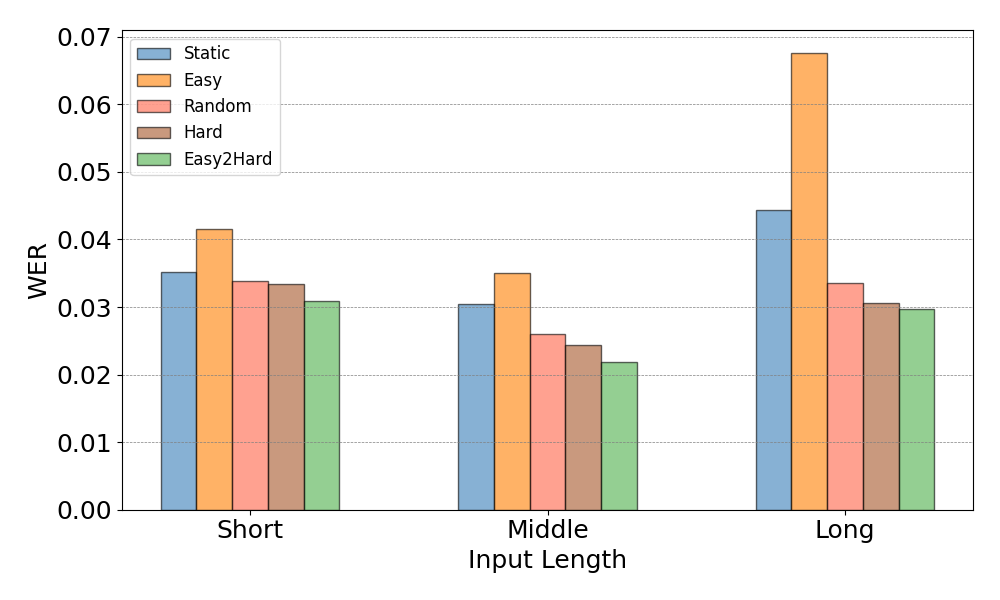}
    \vskip -0.1in
    \caption{Comparing instance-wise pruning strategies across three subsets of the Librispeech "test-clean" set, with instance-wise kept ratio of 50\% for each method.}
    \label{fig:error_analysis}
    \vskip -0.2in
\end{figure}

\vskip -0.05in
\label{error_analysis}
To assess how the presented models affect performance across instances of varying input lengths. We divide the test samples in the ``test-clean'' set of Librispeech into three groups with different input duration, namely, \textit{Short} (0\,-\,8 seconds), \textit{Middle} (8\,-\,16 seconds) and \textit{Long} ($\geq$ 16seconds), respectively. The performance of each group for the \code{Easy}, \code{Random}, \code{Hard} and \code{Easy2hard} methods is presented in Figure~\ref{fig:error_analysis}. Interestingly, we observe that models prioritizing easy instances tend to underperform, especially on longer instances, whereas models that focus on challenging instances show better performance on shorter ones. Overall, the proposed \code{Easy2hard} approach consistently outshines the other methods across all groups.

\section{Conclusion}

In this work, we conduct detailed analysis of dynamic data pruning for ASR, focusing on both instance-wise and time-wise pruning techniques. We demonstrate that these methods can be synergistically employed to maintain performance while achieving significant speed improvements. Among pruning methods, our proposed \code{Easy2hard} method has been found to be the most effective in speech recognition benchmarks. Notably, we observe that pruning up to 30\% of instances, coupled with a 30\% chunk dropping rate, can maintain performance compared to training with the full dataset. Moreover, our findings reveal that time-wise pruning significantly boosts model resilience to lower sampling rates, making it a valuable adjunct to time masking.

\section{Acknowledgements}
This research is part of the research program ‘MegaMind - Measuring, Gathering, Mining and Integrating Data for Self-management in the Edge of the Electricity System’, (partly) financed by the Dutch Research Council (NWO) through the Perspectief program under number P19-25. Additionally, note that only non-Meta authors utilized and processed the datasets (and no dataset pre-processing or processing took place on Meta's servers or facilities). Shiwei Liu is supported by the Royal Society with the Newton International Fellowship. 



\section{References}
\begingroup
\printbibliography[heading=none]
\endgroup

\end{document}